\documentclass[10pt,twocolumn,letterpaper]{article}

\usepackage{preprint}
\usepackage{booktabs}
\usepackage{amssymb}
\usepackage{amsmath}
\usepackage{natbib}
\usepackage{colortbl}
\usepackage{multirow}
\usepackage{multicol}
\usepackage{stfloats}

\definecolor{linkblue}{rgb}{0.21,0.49,0.74}
\usepackage[breaklinks,colorlinks,allcolors=linkblue]{hyperref}
\hypersetup{pdftitle={LGCount: Layout-Guided Diffusion for Zero-Shot Object Counting}}

\title{LGCount: Layout-Guided Diffusion for Zero-Shot Object Counting}

\author{Adriano D'Alessandro \qquad Ali Mahdavi-Amiri \qquad Ghassan Hamarneh\\
Simon Fraser University, Burnaby, Canada\\
{\tt\small \{acdaless, amahdavi, hamarneh\}@sfu.ca}
}

\begin{document}

\maketitle
\begin{figure*}[b]
  \centering

  \includegraphics[width=0.95\textwidth]{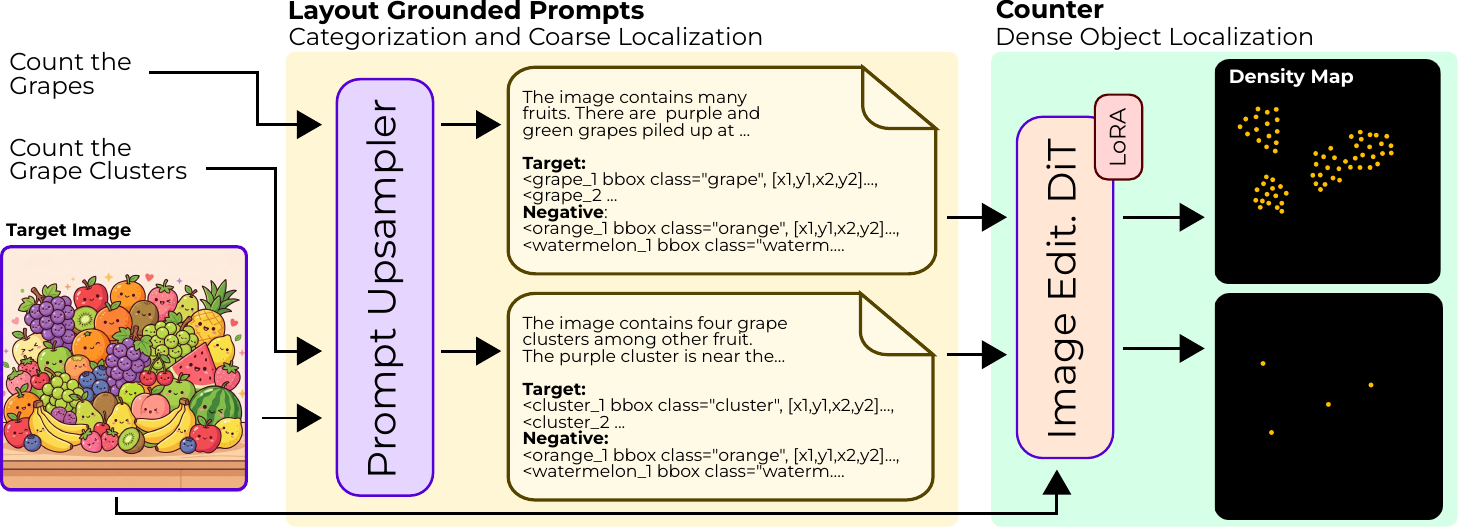}
\caption{Unlike conventional few-shot counters that require manual visual exemplars, our method decomposes counting into a two-stage problem. A VLM-based Prompt Upsampler performs global scene reasoning and coarse localization, converting a query into a \textit{layout} with explicit spatial bounding boxes for positive targets and distractor classes. An image editing diffusion model, adapted via LoRA, then follows these instructions to estimate a Density Map. By decoupling coarse semantic planning from dense pixel-level localization, our approach counts the same photograph as grapes or as grape clusters, with only the request changed.}
  \label{fig:intro_method}
\end{figure*}

\begin{abstract}

Class-agnostic counting requires the user to specify what to count, usually with a class name or a few annotated boxes. We propose a richer text-based specification: a structured layout that names the target, the region that holds it, the distractors to ignore. A prompt upsampler writes this layout from the image and the requested category alone, and an image-editing diffusion model, adapted into a single-step counter, turns it into a density map. Because counting datasets pair each image with a single valid density map, a counter trained on it learns to ignore the layout. We therefore train on synthetic counterfactuals in which one image admits several valid density maps, so that the layout cannot be ignored. On FSC-147 we improve on the previous zero-shot state of the art and we set the best reported result on LOOKALIKES while being told the least about each image. A single photograph can be counted by category, by region, or by granularity without retraining. \href{https://github.com/adrian-dalessandro/LGCount}{Code here.}
\end{abstract}

\section{Introduction}
\label{sec:intro}
Class-agnostic object counting exists as a general-purpose solution for counting objects in dense scenes without any task-specific retraining. Because these methods allow for users to specify categories on-the-fly, a method must identify what qualifies as an object before it can locate and count those objects. Few-shot counters solve this by having a user manually annotate boxes on the image, which provides valuable in-distribution context but does not scale. Few-shot methods are also ambiguous about what should be included and excluded. If a user annotates a man wearing a baseball cap, should the model count hats, people wearing hats, only men, or some other combination? Zero-shot counters resolve some of this ambiguity by naming the category instead, which scales. However, a name is often insufficient to separate fine-grained categories with subtle visual attributes and provides minimal additional context to the model. Recent zero-shot methods narrow the gap by mining pseudo-exemplars using text prompts, but this uses text only as a weak retrieval mechanism for a few-shot pipeline rather than taking advantage of the structured nature of language.
\begin{figure*}[t]
  \centering

  \includegraphics[width=0.94\textwidth]{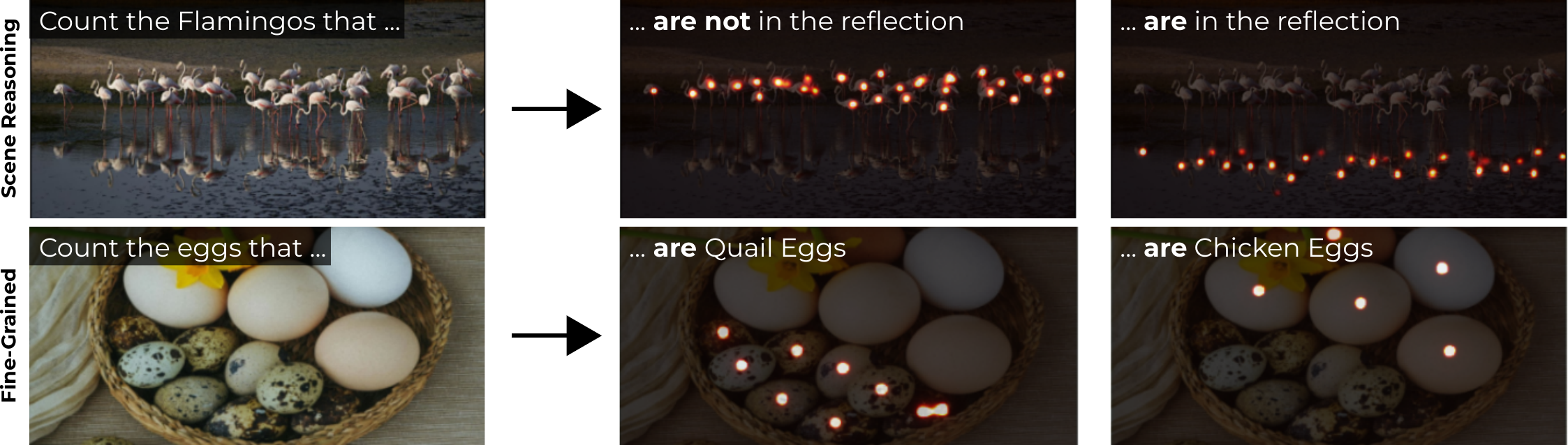}
  \caption{(Top) Our method can isolate categories based on arbitrary attributes, such as whether it appears in a reflection. (Bottom) And it can separate fine-grained categories. Here, we use QWEN3.8 as the upsampler.}
  \label{fig:splash}
\end{figure*}
We view pseudo-exemplar mining as a simpler version of automatic scene comprehension. A pseudo-exemplar provides visual grounding to the model by selecting a few objects within an image, but it sits within a broader set of question: what is in this image, where is each relevant object, what are the distractors, and how do they relate? A method that answers the larger question gets the smaller one for free. It also expresses what a box cannot: which of two lookalikes is meant, which region, which objects to ignore, and what counts as one object. We argue that vision-language models (VLMs)~\cite{liu2023visual, yang2025qwen3} answer the larger question, by providing scene descriptions, identifying objects, and providing coarse localization of scene elements. Given an image and a class name, a VLM will identify the regions that hold the target and point out the distractors without requiring additional user input. The most recent VLMs are capable of producing all of this as a text-based layout, which is a list of named objects with descriptions and coordinates. We use that layout as a rich specification that adapts to each individual image, in which pseudo-exemplars are only one component. However, a layout only specifies what should be counted, and VLMs do not produce accurate counts. We defer that job to a downstream counter.

While the field has coalesced around detection backbones~\cite{groundingdinoeccv,liu2025countse, AminiNaieni24}, these are not designed to consume a layout. We look instead to image diffusion models~\cite{flux-2-2025, podell2024sdxl}, which have proven effective backbones for dense tasks like depth estimation~\cite{pavlovic2026marigoldv2}, and which can generate continuous output that do not require instances to be individually separable. Early such models comprehended layouts poorly, but recent generators separate comprehension from synthesis: a vision-language model, the prompt upsampler, writes a structured layout, and the generator follows it. This is precisely the property we are after. Rather than teaching a detection-based counter to follow instructions, we take a model that already follows instructions and teach it to count. As shown in Figure~\ref{fig:intro_method}, we train a single-step counter that reads the layout and predicts a density map.

Training the backbone to follow instructions and produce accurate density maps requires data in which the instruction matters. FSC-147 provides a single category per image, so every prompt has the same answer and the model can satisfy its loss by ignoring the text. We therefore build synthetic counterfactual scenes that pair one image with several instructions, each with its own density map. A model that ignores the prompt cannot fit such data, and the result resolves distinctions that no exemplar or class name can express. In Figure~\ref{fig:splash}, asked for the birds reflected in the water and then for the real birds, the model returns a different density map for each. Overall, our contributions are:
\begin{itemize}\itemsep2pt
  \item We separate scene understanding from counting. A prompt upsampler writes a structured layout and an image editing diffusion model, adapted into a single-step counter, turns that layout into a density map.
  \item We construct synthetic counterfactuals, scenes in which one image admits several different targets, so that the counter cannot ignore the layout during training.
  \item We report strong zero-shot results on FSC-147 and LOOKALIKES, and show that a single image can be counted by category, by region, or by granularity.
\end{itemize}
\section{Prior Work}
\label{sec:prior_work}
Diffusion models are trained on billions of image and prompt pairs, and the resulting semantic priors have been applied to counting in two related ways. CountDiff~\cite{cact2i2024} learns an object embedding in the text embedding space of a frozen Stable Diffusion model, so that its cross-attention maps highlight the target. T2ICount~\cite{qian2025t2icount} works from a single denoising step, but observes that doing so weakens prompt sensitivity, and recovers it by refining multi-scale text-image features. Both treat the backbone as a feature extractor where the prompt is a class name, features come from attention layers, and a head produces a density map. We instead seek to keep the generative model's conditioning format, which is a grounded layout. Further related work appears in Appendix~\ref{sec:related}.
\begin{figure*}[t]
  \centering

  \includegraphics[width=\textwidth]{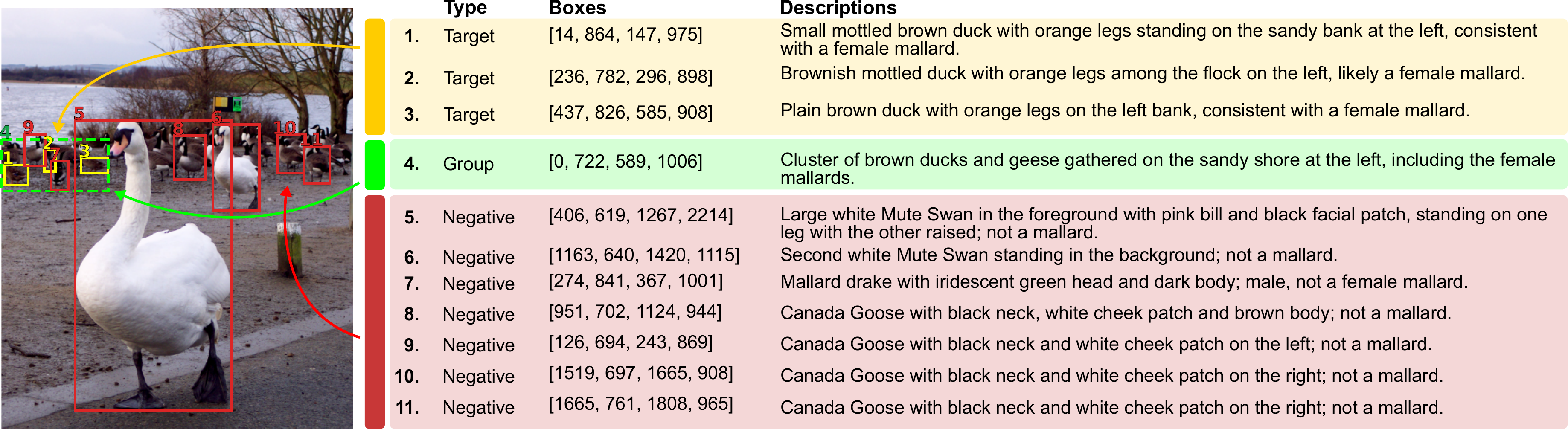}
\caption{\textbf{A layout.} Given the image and the request \emph{female mallard}, the upsampler returns rows of three kinds: instances of the target, one group row bounding the region that holds them, and negatives, which it proposes itself rather than being told. Every row carries a box and a description, and the layout reaches the counter as text.}
  \label{fig:layouts}
\end{figure*}
\section{Methodology}\label{sec:methodology}
Zero-shot class-agnostic counting methods are typically conditioned on a class name, which is passed through a language model, resulting in a fixed text embedding. This leaves the counting backbone to perform two different tasks: it must decide which objects match the fixed embedding, and enumerate all of them. However, class names provide minimal information about what the object actually looks like and how it might be presented in a given scene.

More recent methods have adopted the strategy of first selecting high-confidence pseudo-exemplars using text guidance, then counting them in a few-shot setting~\cite{AminiNaieni24, liu2025countse}. Effectively, this is a two-pass arrangement, in which a text-conditioned prediction is fed back to the model as its own exemplars. This attempts to ground the class name in the image, but still requires a fixed model to isolate the correct exemplars in multi-category scenes, and it still leaves the counter to interpret how those exemplars relate to the semantics of the scene. This raises a question: why should the counter be responsible for deciding what to count? Counters have no mechanism for reasoning about scenes, and a semantic failure can only be repaired by retraining.

Vision-language models, by contrast, can interpret a user query, reason about a scene, propose exemplars and negative categories, describe the objects involved, and write all of it down as a structured layout. When used in this setting, to enrich a user query, we refer to them as \textit{prompt upsamplers}. What they do poorly is count objects in dense scenes, and they are inefficient as detectors. We therefore propose giving each role to the model most suited to it: a prompt upsampler reasons about the scene semantics and writes a structured layout describing where the objects are, and a fixed counter produces a count that follows this layout, without deciding for itself which objects belong. Because the layout arrives as text at inference, the counter is trained once, and any vision-language model able to reason about a scene and write a layout can be substituted as the prompt upsampler without retraining the counter. Our method is built around this division of responsibilities.

\subsection{Dividing Responsibilities}\label{sec:dividing}
To divide responsibilities, we take advantage of a trend in image generation where diffusion models are prompted by first performing visual planning before producing structured layouts~\cite{lian2024llmgrounded, feng2023layoutgpt, yang2024mastering, flux-3-image-klein-2026}. This provides us an opportunity, as prior work has also demonstrated that T2I diffusion models can perform dense tasks well~\cite{pavlovic2026marigoldv2}, and be modified into object counters~\cite{qian2025t2icount, cact2i2024}. In the following section, we overview the layout templates that facilitate communication between the upsampler and counter, and how we train the counter.

\begin{figure*}[t]
  \centering

  \includegraphics[width=\textwidth]{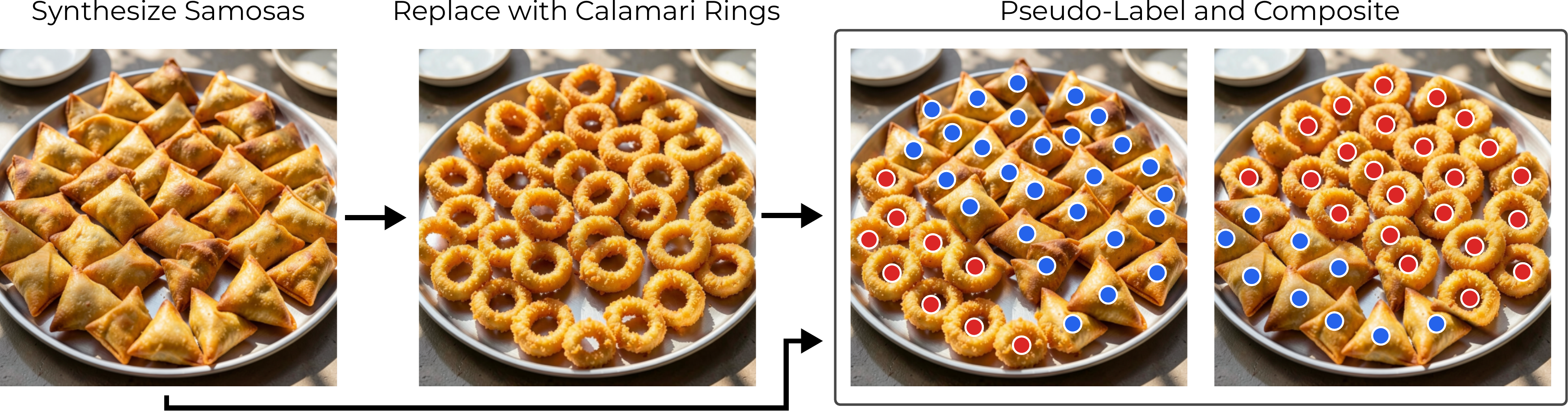}
    \caption{\textbf{Building a synthetic counterfactual.} We synthesize a scene containing one category using a text-to-image diffusion model, then use an image-to-image editing model to replace the target objects, and only those objects, with those of a different category. We then pseudo-label both images independently, and composite them. Because the two halves are images of the same scene, the composite differs in what the objects are and in nothing else, and the image supports a different density map under each request.}
  \label{fig:make_composite}
\end{figure*}
\subsubsection{The Layout Specification}\label{sec:layout}
A layout is a text prompt that communicates a set of regions to the counter. Each row carries an identifier, bounding box coordinates, a free-text description of what the object looks like, and a declared role. A \emph{group} row tells the counter to count inside a specific area rather than everywhere. \emph{Target} rows mark individual instances that are examples of the target, so the counter is shown what one looks like rather than being supplied only a name. \emph{Negative} rows mark distractors and tell the counter what to ignore.

Figure~\ref{fig:layouts} shows how a layout maps onto an image for a single request, as written by the upsampler when asked for a ``female mallard''. It marks three female mallards as targets, one group region holding the flock, and seven negatives. That string, and nothing else, is what the counter receives. Rows compose freely. A layout may place instances and negatives inside a group region, as here; give a group region alone; or give targets with no region at all. Nothing in the format privileges any specific type of row, and the same scene admits many different layouts under different requests. A layout therefore selects along three axes at once. It dictates which objects, isolating the female mallards from other waterfowl. It dictates which region, allowing requests like counting only horses reflected in a pond rather than the physical horses on the bank. And it dictates at what granularity, since a row bounding a whole object asks for whole objects even when the image contains smaller countable parts.

We did not invent this format. The generative backbone is already capable of editing images using a grounded layout before we apply any counting supervision, so we re-express the upsampler's detections into that convention rather than defining our own. Because this layout acts as the counter's only conditioning channel and arrives as plain text, either stage can be replaced without retraining the other.

\subsubsection{The Counter}
The counter is an image editing diffusion model, chosen because layouts are its native conditioning format. We train it to consume a layout and emit a density map, a continuous dense field that localises the requested objects and integrates to their count. We adapt it with a low-rank adapter and otherwise leave the backbone frozen, and we treat it as a single-step regressor rather than a generative sampler, following~\cite{pavlovic2026marigoldv2}: one forward pass, no sampler, no schedule and no seed, so a prediction is deterministic.

The difficulty is that this ability, which the backbone already has, is easily trained away. Standard counting datasets such as FSC-147~\cite{Ranjan_2021_CVPR} provide dense annotations across a variety of scenes and categories, but annotate one category per image, and many images contain only a single kind of countable object. The resulting objective can be satisfied without reading the layout at all, by counting whichever object repeats. The risk is not that the counter cannot follow a layout, but that counting supervision teaches it not to.

\begin{figure*}[t]
  \centering

  \includegraphics[width=\textwidth]{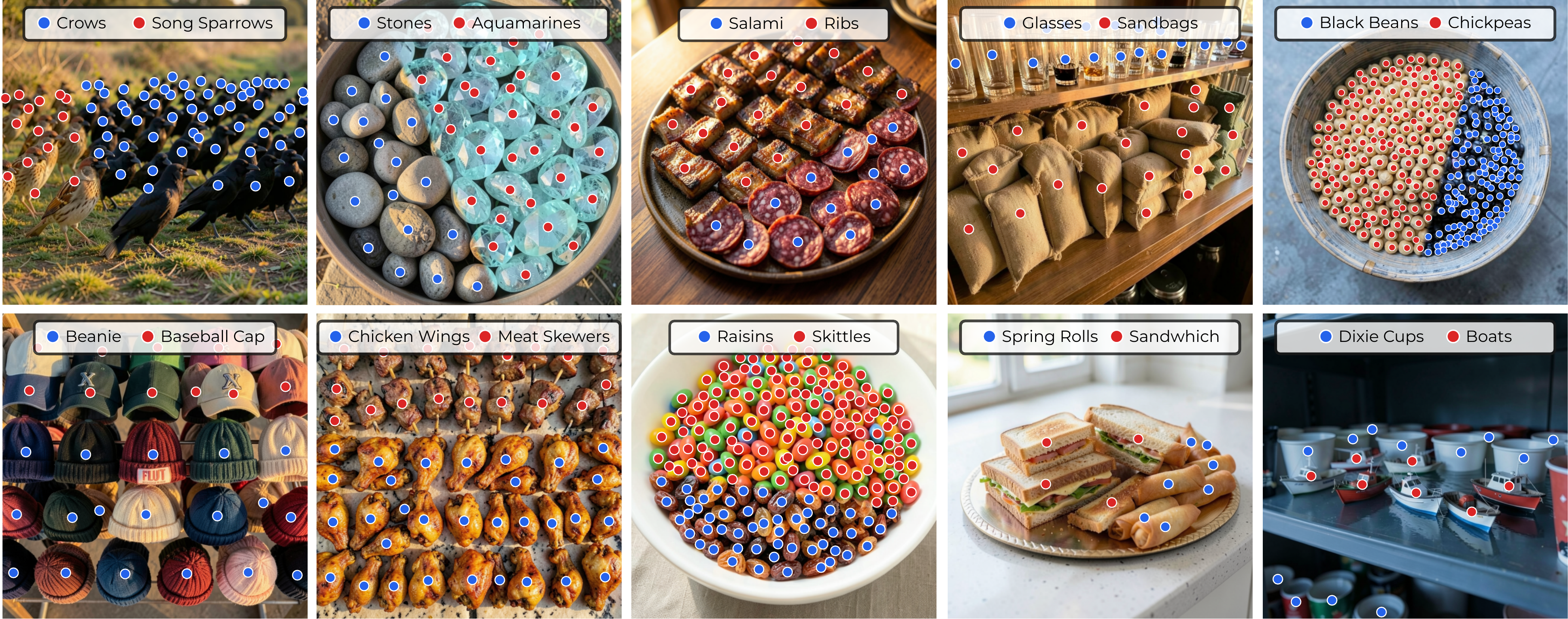}
  \caption{\textbf{Synthetic counterfactuals.} Each scene pairs two categories with their pseudo-labelled points. The pairs span a range of visual similarity, from categories separable by shape and colour to siblings that differ only in surface detail.}
  \label{fig:composite_examples}
\end{figure*}
\subsection{Generating Synthetic Counterfactuals}\label{sec:counterfactuals}
To make the layout load-bearing, training needs images that admit more than one valid density map, selected by the prompt. Using negative prompts with empty density maps does not achieve this, as the objective is satisfied by a gain
control, a switch between predicting nothing and predicting the one valid map. We therefore construct counterfactual images, which contain two categories at once and so require a different density map for each.

The simplest construction composites two real photographs along a shared seam.
This gives real pixels, human annotations on both sides, and arbitrary pairing
of categories, and we use it as our basic augmentation. It has two limitations.
First, the two categories are separated by a visible seam, an appearance cue at
the boundary that is predictive of the target but unrelated to the objects
themselves. Second, existing counting datasets have few object categories. For
example, FSC-147 has 89 training categories.

We therefore propose a strategy for generating synthetic counterfactuals that removes this shortcut and expands the number of categories available. We generate an image of a single category with a text-to-image diffusion model, then use an image-to-image model to replace the objects, and only the objects, with those of a sibling category, and composite the two (Figure~\ref{fig:make_composite}). Both halves are renderings of the same scene, so they differ in what the objects are and in nothing else: same container, same lighting, same arrangement, same background. There is therefore no appearance cue at the boundary beyond the objects themselves.

Both renderings are pseudo-labelled independently by an off-the-shelf counting model, which is reliable here because each image contains a single category by design. The annotated points on the composite are then evaluated with a vision model to identify objects that were not successfully replaced, and those points are reassigned to the original category, or the image is discarded if too few objects were replaced. The resulting pool spans several hundred category pairs (Figure~\ref{fig:composite_examples}), from pairs differing in shape and colour to siblings separable only by texture.

\subsection{Prompt Construction}\label{sec:prompts}
Training layouts are constructed from each annotated record, while at inference they are re-expressed from the upsampler's output. A subset of objects is selected as target rows, and the prompt instructs the counter to locate the remaining objects. How many instances are cited, whether a region or negatives appear, the row order, and the phrasing of the instruction are all sampled, so the counter cannot satisfy the objective by keying on a constant string.

Negatives come from the second annotated category where a record has one, and otherwise from the upsampler, which volunteers them unasked. They are also corrupted deliberately, with a small probability of dropping one or replacing it with a category that is not present. At inference, the table comes from a model that will sometimes be wrong, and a counter trained only on perfect specifications treats them as authoritative.

\subsection{Learning from Density Maps}\label{sec:density}
Despite the trend towards using detection backbones for class-agnostic counting~\cite{AminiNaieni24, Amini_Naieni_2026_CVPR}, density maps are better suited to counting with a generative image backbone, which is able to represent dense continuous spatial structures. Point annotations are rendered into a density map using a Gaussian kernel, so that the map integrates exactly to the object count.

Three losses supervise the model during training. The first is a regression
term in the latent space,
\begin{equation}
\mathcal{L}_{\mathrm{lat}} = \big\lVert\, \hat{z} - \mathcal{E}(y) \,\big\rVert_2^2 ,
\end{equation}
between the predicted density latent $\hat{z}$ and the ground-truth density map $y$ passed through the frozen encoder $\mathcal{E}$, following prior
work~\cite{pavlovic2026marigoldv2}. Because the prediction is compared against the target at corresponding locations, every annotated point becomes a hard constraint. A substantial part of our data comes from the synthetic composites, whose points are provided by a noisy detector rather than by a person, and under this objective a model must reproduce a spurious detection.

The second term therefore measures position by transport rather than by pixel overlap, following the crowd counting literature~\cite{lin2023optimal, wan2021generalized}. Writing $a$ for the decoded prediction $\hat{y}$ as a measure over pixels, $b$ for unit mass at each annotated point, and $C_{ij}$ for the squared distance between them,
\begin{equation}
\mathcal{L}_{\mathrm{GL}} = \min_{P \ge 0}\ \langle C, P\rangle
  - \varepsilon H(P) + \tau\,\mathcal{D}(P, a, b),
\end{equation}
with $\mathcal{D} = \lVert P\mathbf{1} - a\rVert_2^2 +
\lVert P^{\!\top}\mathbf{1} - b\rVert_1$ and $H$ the entropy of the transport plan. We adopt this formulation for two reasons. Positional error costs transport distance rather than pixel mismatch, so a prediction a few pixels off is cheap, which is the right response to pseudo-labeler jitter and to the genuine ambiguity of where a blob belongs. And the problem is unbalanced, so unmatched mass is paid for at a bounded price rather than forcing a blob into existence.

The third term scores the predicted mass that falls near the points of a declared negative,
\begin{equation}
\mathcal{L}_{\mathrm{neg}} = \frac{\sum_x \hat{y}(x)\,M(x)}{\sum_x \hat{y}(x) + \epsilon},
\end{equation}
where $M$ marks pixels close to a negative point. Neither of the other terms charges for this. The transport term is computed against the positive points only, and the latent regression is dominated by the foreground, so a region the layout declares a negative is supervised only by its target happening to be zero there. Because the term is a fraction of predicted mass rather than an absolute quantity, it redistributes mass instead of competing for the total, and at the optimum it is zero and contributes no gradient. It applies only to records that carry a second annotated category.
\begin{table*}[t]
\centering
\begin{tabular}{lccccccc}
\toprule
Method & Year & Backbone & Prompt & Pseudo & Synthetic & MAE & RMSE  \\
\midrule
FamNet~\cite{Ranjan_2021_CVPR} & 2021 & ResNet & Visual & -- & -- & 22.08 & 99.54 \\
CACViT~\cite{wang2024vision} & 2024 & ViT & Visual &-- & -- & 9.13 & 48.96 \\
DAVE~\cite{Pelhan_2024_CVPR} & 2024 & ResNet & Visual & -- & -- & \textbf{8.66} & \textbf{32.36} \\
\midrule
CountGD & 2024 & G.Dino & Vis. + Text & -- &-- & \textbf{\underline{5.74}} & 24.09 \\
CountGD++& 2026 & G.Dino & Vis. + Text & -- & -- & 5.86 & \textbf{\underline{18.79}} \\
\midrule 
GroundingDINO~\cite{groundingdinoeccv} & 2024 & G.Dino & Text & $\times$ & $\times$ & 11.52 & 97.37 \\
CountGD~\cite{AminiNaieni24} & 2024 & G.Dino & Text & $\times$ & $\times$ & 12.98 & 98.35 \\
GroundingREC~\cite{Dai_2024_CVPR} & 2024 & G.Dino & Text & $\times$ & $\times$ & 10.12 & 107.19 \\
CountDiff~\cite{cact2i2024} & 2024 & Diffusion & Text & $\times$ & $\times$ & 14.83 & 103.15 \\
T2ICount~\cite{qian2025t2icount} & 2025 & Diffusion & Text & $\times$ & $\times$ & 11.76 &  97.86 \\
CountSE~\cite{liu2025countse} & 2025 & G.Dino & Text & $\checkmark$ & $\times$ & 7.84 & 82.99 \\
CountGD++~\cite{Amini_Naieni_2026_CVPR} & 2026 & G.Dino & Text & $\times$ & $\times$ & 16.55 & 129.76 \\
CountGD++~\cite{Amini_Naieni_2026_CVPR} & 2026 & G.Dino & Text & $\checkmark$ & $\times$ & 10.29 & 33.52 \\
CountGD++~\cite{Amini_Naieni_2026_CVPR} & 2026 & G.Dino & Text & $\checkmark$ & $\checkmark$ & 8.39 & 27.03 \\
\midrule
\rowcolor{gray!10} LGCount (ours) & -- & Diffusion & Text/Layout & $\checkmark$ & $\times$ & \textbf{7.06}& \textbf{19.38} \\
\bottomrule
\end{tabular}
\caption{Comparison with prior work on the FSC-147 dataset. Methods are grouped by what they are given at test time: visual exemplars, visual exemplars with text, or text alone. \emph{Pseudo} marks methods that mine pseudo-exemplars from the text prompt, and \emph{Synthetic} marks those that use synthetic visual exemplars. Bold marks best-in-group and underline marks best-overall.}
\label{tab:main_results}
\end{table*}

\subsection{Inference}\label{sec:inference}
At inference the counter is a single-step regressor. The image is encoded once and the density latent obtained in one forward pass at a fixed timestep.
There is no sampler, schedule or seed, so the prediction is deterministic. The latent is decoded and the count read as the sum of the density map. 

A single forward pass is calibrated only over a bounded range. The counter sees a fixed number of tokens regardless of object density, so the binding quantity is objects per token. Tiling raises the token budget per image region: each pass covers less of the scene with the same number of tokens, bringing the count within the pass back inside the calibrated range. We additionally pass non-empty patches
through a generative upscaler, following prior work~\cite{Amini_Naieni_2026_CVPR}, which supplies detail so that the additional tokens have something to separate. This leaves the question of when to tile, and the counter cannot answer it since a single pass is precisely what is unreliable.
\begin{figure*}[!tp]
  \centering

  \includegraphics[width=0.94\textwidth]{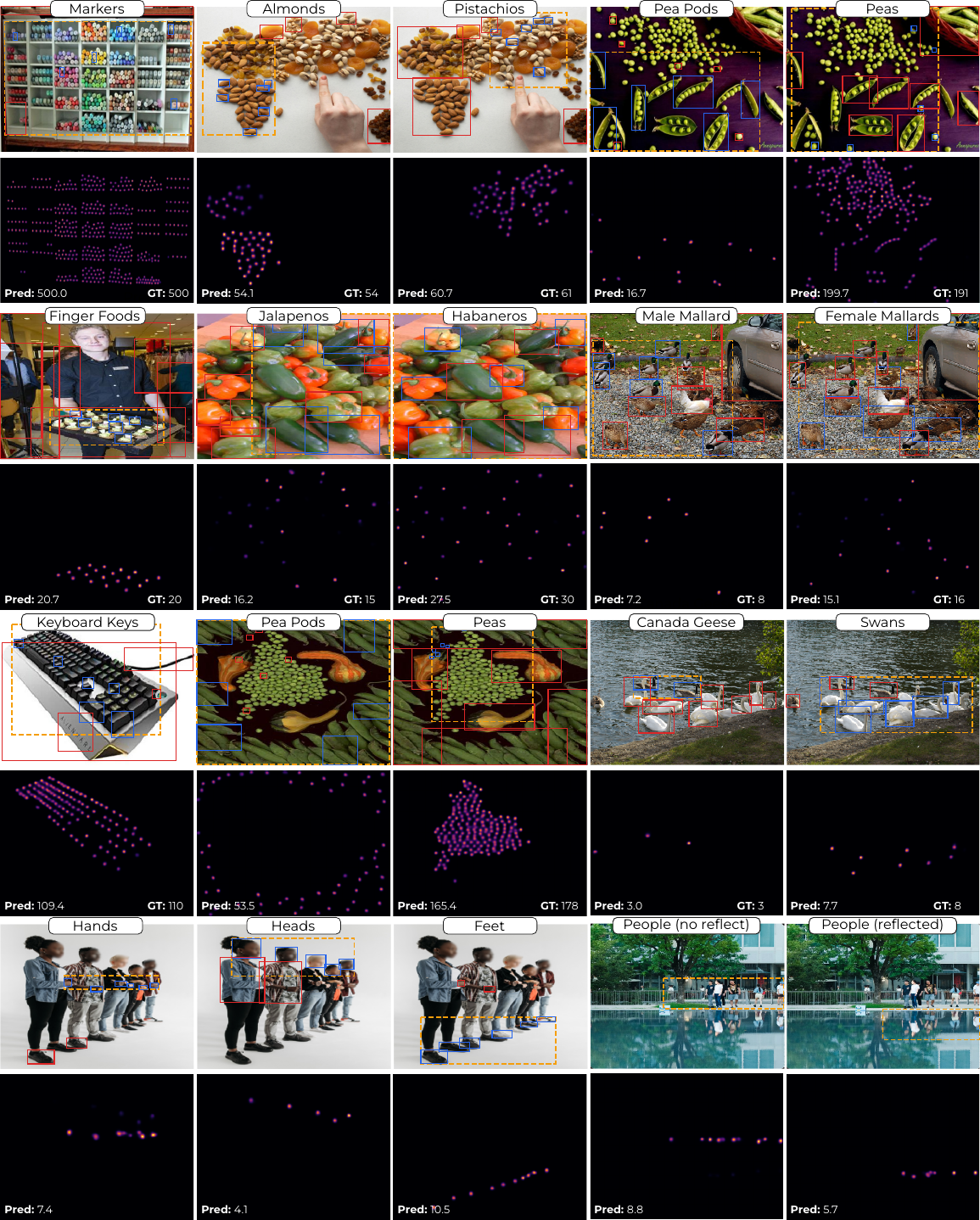}
\caption{\textbf{Qualitative results.} Each pair of rows shows the input with its layout overlaid (amber = group, blue = targets, red = negatives) and the resulting density map. Every panel comes from the same counter and weights, with only the layout changing. Counts span two orders of magnitude, from a few birds to several hundred markers. Fine-grained pairs separate semantically related objects such as jalapenos against habaneros, or male against female mallards. Others show whole-versus-part control, such as pea pods against the loose peas spilled around them, and human body parts. The last pair is a structural request, counting the people beside a pool and then their reflections. Ground truth is reported only where it ships with a dataset.}
  \label{fig:qualitative}
\end{figure*}
We return to the upsampler for this. Vision-language models approximate counts well in sparse scenes and while their precision degrades sharply as scenes become crowded, they continue to identify the correct order of magnitude, which is all the decision requires. We use the upsampler's count to decide when to tile, and then treat the untiled count, the tiled count and the upsampler's count as three estimates and report their median.

\section{Experiments}\label{sec:experiments}

\begin{table}[t]
\caption{Results on LOOKALIKES, a fine-grained benchmark. \emph{Neg}: the method is told, per image, which lookalike categories are present. \emph{Infer.}: zero-shot (ZS) or test-time adapted (TTA).}
  \label{tab:lookalikes}
  \centering
  \setlength{\tabcolsep}{3.0pt}
  \begin{tabular}{l l c c cc}
    \toprule

    Method & Ref. & Neg & Infer. & MAE$\downarrow$ & RMSE$\downarrow$ \\
    \midrule
    CountGD \footnotesize{+FiGO}~\cite{dalessandro2025justsaytheword} & Text & $\checkmark$ & TTA & \textbf{13.31} & \textbf{23.36} \\
    GrREC \footnotesize{+FiGO}~\cite{dalessandro2025justsaytheword} & Text &  $\checkmark$ & TTA & 16.15 & 26.99 \\
    \midrule
    CountGD++~\cite{Amini_Naieni_2026_CVPR} & Text & $\checkmark$ & ZS & 24.6 & 37.3 \\
    CountGD++~\cite{Amini_Naieni_2026_CVPR} & Text\tiny{\textsc{+PS}} & $\checkmark$ & ZS & 29.1 & 45.0 \\
    RECOUNT~\cite{d2026recount} & Image & $\checkmark$ & ZS & \textbf{8.41} & \textbf{14.67} \\
    \midrule
    DAVE~\cite{Pelhan_2024_CVPR} & Text & $\times$ & ZS & 53.10 & 64.44 \\
    CountGD~\cite{AminiNaieni24} & Text & $\times$ & ZS & 30.17 & 45.01 \\
    GrREC~\cite{Dai_2024_CVPR} & Text & $\times$ & ZS & 18.51 & 30.39 \\
    \midrule
    \rowcolor{gray!10} LGCount (Ours) & Text & $\times$ & ZS & \textbf{\underline{7.97}} & \textbf{\underline{14.34}} \\
    \bottomrule
  \end{tabular}
\end{table}
\paragraph{Datasets.} We evaluate on FSC-147~\cite{Ranjan_2021_CVPR}, the standard class-agnostic counting benchmark, and on LOOKALIKES~\cite{dalessandro2025justsaytheword}, which tests fine-grained discrimination. FSC-147 contains 6{,}135 images over 147 categories, each annotated with points for a single category. LOOKALIKES instead contains scenes in which several visually similar categories appear at once, with 37 fine-grained subcategories. We report the macro average over subcategories.
\paragraph{Implementation.} The counter is FLUX.3 Image
Klein~\cite{flux-3-image-klein-2026}, adapted with a rank-128 LoRA. The prompt upsampler is Qwen3.8-27B~\cite{qwen38}. We defer additional implementation details to Appendix~\ref{sec:supp_impl}.
\subsection{Results}\label{sec:results}
On FSC-147 (Table~\ref{tab:main_results}) the strongest text-only results all use pseudo-exemplars, with CountSE at $7.84$ and CountGD++ at $8.39$. Ours are proposed in context by the upsampler and arrive as instance rows inside the layout, rather than coming from a first pass of the counter over its own output. This structured layout is a richer source of information, and our
method improves over CountSE by $10.0\%$. Text-only counting has closed the gap to few-shot, and the remaining gap is to methods given both exemplars and text, where CountGD reaches $5.74$ against our $7.06$. Ours is also the only recent diffusion-based method to be competitive with detector-based ones.

On LOOKALIKES (Table~\ref{tab:lookalikes}) we obtain the best results while
being given the least information. The methods marked \emph{Neg} are told, per
image, which lookalike categories are present and must be separated from the
request. We are given only the image and the positive requested category, and
the upsampler proposes the distractors itself. The closest method, RECOUNT, is
queried with a reference image rather than text. Among methods that are
text-queried, zero-shot, and not supplied with the distractors, the strongest
prior result is GrREC at $18.51$, which we more than halve.

\paragraph{Qualitative} Figure~\ref{fig:qualitative} shows predictions from our
model given different layouts produced by QWEN3.8~\cite{qwen38}. The
fine-grained pairs are the clearest evidence that the counter follows the
specification. One photograph of mixed nuts yields $54.1$ when the request
names almonds and $60.7$ when it names pistachios, against ground truths of
$54$ and $61$, and the same holds for the other fine-grained pairs in the
figure. In each pair the photograph is identical and only the request, and the layout it produces, differ.

Granularity is controlled the same way. Asked for pea pods the counter places
mass on the pods, and asked for peas on the same image it places mass on the
loose peas around them. Body-part cells repeat this on one photograph of a
group of people, where feet, hands and heads are each requested in turn. This
is an ambiguity that an exemplar image expresses poorly, since a photograph of a pea pod is equally a photograph of the peas inside it and only the text can
say which is meant. The final pair asks for a region rather than a category,
requesting the people beside a pool and then the people reflected in it.

\subsection{Ablations}

\begin{table}[t]
\caption{The upsampler alone against the full system. The counter contributes
most where the upsampler is weakest, in dense scenes.}
  \label{tab:vlm_compare}
  \centering
  \setlength{\tabcolsep}{4.5pt}
  \begin{tabular}{l c c c c}
    \toprule
    \multirow{2}{*}{Method} & \multicolumn{2}{c}{FSC147} & \multicolumn{2}{c}{LookAlikes} \\
    \cmidrule(lr){2-3} \cmidrule(lr){4-5}
    & MAE$\downarrow$  & RMSE$\downarrow$ & MAE$\downarrow$ & RMSE$\downarrow$ \\
    \midrule
    Qwen3.8 & 13.48 & 67.97 & 8.80 & 19.14\\
    LGCount\footnotesize{+Qwen3.8} & \textbf{7.06}& \textbf{19.38} & \textbf{7.97} & \textbf{14.34}\\
    \bottomrule
  \end{tabular}
\end{table}
\begin{table}[t]
  \caption{Ablating synthetic counterfactuals. $\Delta$ is the cost of removing them.}
  \label{tab:no_composites}
  \centering
  \setlength{\tabcolsep}{3.5pt}
  \begin{tabular}{r c c c c}
    \toprule
    \multirow{2}{*}{Method} & \multicolumn{2}{c}{FSC147} & \multicolumn{2}{c}{LookAlikes} \\
    \cmidrule(lr){2-3} \cmidrule(lr){4-5}
    & MAE$\downarrow$  & RMSE$\downarrow$ & MAE$\downarrow$ & RMSE$\downarrow$ \\
    \midrule
    FSC-147 Only & 9.57 & 28.98 & 12.68 & 22.89\\
    ~ + \footnotesize{Counterfactuals} & \textbf{7.06}& \textbf{19.38} & \textbf{7.97} & \textbf{14.34}\\
    \midrule
    $\Delta$ & +2.51 & +9.60 & +4.71 & +8.55\\
    \bottomrule
  \end{tabular}
\end{table}
\paragraph{Contribution of the counter} The upsampler alone is already a reasonable counter where scenes are sparse and the difficulty is semantic, reaching $8.80$ on LOOKALIKES, but it degrades badly as scenes become dense, with an RMSE of $67.97$ on FSC-147. Adding the counter nearly halves MAE on FSC-147 and cuts its RMSE more than threefold, while the gain on LOOKALIKES is smaller, which is what the division of labour predicts: the upsampler supplies the semantics and the counter supplies the magnitude.
\paragraph{Synthetic counterfactuals} Training without them costs $2.51$ MAE on FSC-147 and $4.71$ on LOOKALIKES, the  larger loss falling on the benchmark that requires telling similar categories apart. The cost in RMSE is larger still on
both, which indicates a categorical failure rather than a uniform one. With nothing forcing the layout to be read, the counter loses the ability to parse the layout, and even underperforms QWEN3.8 by itself. 

\section{Conclusion}
We present a counter that does not decide what to count. A prompt upsampler reads the scene and writes a layout naming the target, the region and the objects to ignore, and a counter trained once turns that layout into a density map. To keep the layout load-bearing, we train it on synthetic counterfactuals that preserve the routing ability already present in the underlying diffusion model. The benefit appears where the two abilities pull apart: on FSC-147 the counter supplies the magnitude the upsampler lacks, and on LOOKALIKES the layout supplies a distinction the counter could not make alone. The same mechanism lets one image be counted several ways, by category, by region or by granularity, without retraining.

\section*{LLM Usage Statement}
Claude Code (Anthropic), an AI coding assistant, was used throughout this work. In the writing, it assisted in text revision in all sections, and checked grammar and consistency of terminology. All technical claims, numbers and their interpretation were devised by, written by and verified by the authors. All research ideas were independently produced by the authors. In the research, it was used to assist in writing and debugging data processing, training, evaluation and figure-generation code, and to help analyze experimental results. It was not used to generate experimental results, and no text, number or figure in this paper was accepted without author verification. A vision-language model is additionally a component of the proposed method, as described in Section~\ref{sec:methodology}.
{
    \small
    \bibliographystyle{ieeenat_fullname}
    \bibliography{main}

@misc{flux-3-image-klein-2026,
    author={Black Forest Labs},
    title={{FLUX 3 Image Klein}},
    year={2026},
    howpublished={\url{https://bfl.ai/models/flux-3-image-klein}},
}

@inproceedings{liu2025countse,
  title={CountSE: Soft Exemplar Open-set Object Counting},
  author={Liu, Shuai and Zhang, Peng and Zhang, Shiwei and Ke, Wei},
  booktitle={Proceedings of the IEEE/CVF International Conference on Computer Vision},
  pages={21536--21546},
  year={2025}
}

@InProceedings{Amini_Naieni_2026_CVPR,
    author    = {Amini-Naieni, Niki and Zisserman, Andrew},
    title     = {CountGD++: Generalized Prompting for Open-World Counting},
    booktitle = {Proceedings of the IEEE/CVF Conference on Computer Vision and Pattern Recognition (CVPR)},
    month     = {June},
    year      = {2026},
    pages     = {37725-37734}
}

@InProceedings{AminiNaieni24,
  author = "Amini-Naieni, N. and Han, T. and Zisserman, A.",
  title = "CountGD: Multi-Modal Open-World Counting",
  booktitle = "Advances in Neural Information Processing Systems (NeurIPS)",
  year = "2024",
}

@inproceedings{qian2025t2icount,
               title={T2ICount: Enhancing Cross-modal Understanding for Zero-Shot Counting}, 
               author={Qian, Yifei and Guo, Zhongliang and Deng, Bowen and Lei, Chun Tong and Zhao, Shuai and Lau, Chun Pong and Hong, Xiaopeng and Pound, Michael P},
               year={2025},
               booktitle={Proceedings of the IEEE/CVF conference on computer vision and pattern recognition}
}

@InProceedings{cact2i2024,
author="Hui, Xiaofei
and Wu, Qian
and Rahmani, Hossein
and Liu, Jun",
title="Class-Agnostic Object Counting with Text-to-Image Diffusion Model",
booktitle="Computer Vision -- ECCV 2024",
year="2024",
publisher="Springer Nature Switzerland",
address="Cham",
pages="1--18",
}

@article{pavlovic2026marigoldv2,
    author = {Pavlovic, Igor and Wandel, Thiemo and Obukhov, Anton and Bartolomei, Luca and Davydov, Andrey and Tosi, Fabio and Poggi, Matteo and S{\"u}sstrunk, Sabine and Dai, Dengxin},
    title = {Marigold V2: Revisiting Diffusion Transformers for Monocular Depth Estimation},
    year = {2026},
    issue_date = {December 2026},
    publisher = {Association for Computing Machinery},
    volume = {45},
    number = {6},
    url = {https://doi.org/10.1145/3842528},
    doi = {10.1145/3842528},
    journal = {ACM Trans. Graph.},
    month = dec,
    articleno = {204},
    numpages = {14}
}

@article{
lian2024llmgrounded,
title={{LLM}-grounded Diffusion: Enhancing Prompt Understanding of Text-to-Image Diffusion Models with Large Language Models},
author={Long Lian and Boyi Li and Adam Yala and Trevor Darrell},
journal={Transactions on Machine Learning Research},
issn={2835-8856},
year={2024}
}

@article{feng2023layoutgpt,
  title={Layoutgpt: Compositional visual planning and generation with large language models},
  author={Feng, Weixi and Zhu, Wanrong and Fu, Tsu-jui and Jampani, Varun and Akula, Arjun and He, Xuehai and Basu, Sugato and Wang, Xin Eric and Wang, William Yang},
  journal={Advances in neural information processing systems},
  volume={36},
  pages={18225--18250},
  year={2023}
}

@inproceedings{
yang2024mastering,
title={Mastering Text-to-Image Diffusion: Recaptioning, Planning, and Generating with Multimodal {LLM}s},
author={Ling Yang and Zhaochen Yu and Chenlin Meng and Minkai Xu and Stefano Ermon and Bin CUI},
booktitle={Forty-first International Conference on Machine Learning},
year={2024},
url={https://openreview.net/forum?id=DgLFkAPwuZ}
}

@InProceedings{Ranjan_2021_CVPR,
    author    = {Ranjan, Viresh and Sharma, Udbhav and Nguyen, Thu and Hoai, Minh},
    title     = {Learning To Count Everything},
    booktitle = {Proceedings of the IEEE/CVF Conference on Computer Vision and Pattern Recognition (CVPR)},
    month     = {June},
    year      = {2021},
    pages     = {3394-3403}
}

@article{dalessandro2025justsaytheword,
  title   = {FiGO: Fine-Grained Object Counting without Annotations},
  author  = {D'Alessandro, Adrian and Mahdavi-Amiri, Ali and Hamarneh, Ghassan},
  journal = {arXiv preprint arXiv:2504.11705},
  year    = {2025}
}

@inproceedings{d2024afreeca,
  title={Afreeca: Annotation-free counting for all},
  author={D’Alessandro, Adriano and Mahdavi-Amiri, Ali and Hamarneh, Ghassan},
  booktitle={European Conference on Computer Vision},
  pages={75--91},
  year={2024},
  organization={Springer}
}

@InProceedings{Pelhan_2024_CVPR,
    author    = {Pelhan, Jer and Luke\v{z}i\v{c}, Alan and Zavrtanik, Vitjan and Kristan, Matej},
    title     = {DAVE - A Detect-and-Verify Paradigm for Low-Shot Counting},
    booktitle = {Proceedings of the IEEE/CVF Conference on Computer Vision and Pattern Recognition (CVPR)},
    month     = {June},
    year      = {2024},
    pages     = {23293-23302}
}

@misc{flux-2-2025,
    author={Black Forest Labs},
    title={{FLUX.2: Frontier Visual Intelligence}},
    year={2025},
    howpublished={\url{https://bfl.ai/blog/flux-2}},
}

@InProceedings{Xu_2023_CVPR,
    author    = {Xu, Jingyi and Le, Hieu and Nguyen, Vu and Ranjan, Viresh and Samaras, Dimitris},
    title     = {Zero-Shot Object Counting},
    booktitle = {Proceedings of the IEEE/CVF Conference on Computer Vision and Pattern Recognition (CVPR)},
    month     = {June},
    year      = {2023},
    pages     = {15548-15557}
}

@InProceedings{groundingdinoeccv,
author="Liu, Shilong
and Zeng, Zhaoyang
and Ren, Tianhe
and Li, Feng
and Zhang, Hao
and Yang, Jie
and Jiang, Qing
and Li, Chunyuan
and Yang, Jianwei
and Su, Hang
and Zhu, Jun
and Zhang, Lei",
title="Grounding DINO: Marrying DINO with Grounded Pre-training for Open-Set Object Detection",
booktitle="Computer Vision -- ECCV 2024",
year="2024",
publisher="Springer Nature Switzerland",
address="Cham",
pages="38--55"
}

@InProceedings{Dai_2024_CVPR,
    author    = {Dai, Siyang and Liu, Jun and Cheung, Ngai-Man},
    title     = {Referring Expression Counting},
    booktitle = {Proceedings of the IEEE/CVF Conference on Computer Vision and Pattern Recognition (CVPR)},
    month     = {June},
    year      = {2024},
    pages     = {16985-16995}
}

@inproceedings{wang2019learning,
  title={Learning from Synthetic Data for Crowd Counting in the Wild},
  author={Wang, Qi and Gao, Junyu and Lin, Wei and Yuan, Yuan},
  booktitle={Proceedings of IEEE Conference on Computer Vision and Pattern Recognition (CVPR)},
  pages={8198--8207},
  year={2019}
}

@inproceedings{lin2023optimal,
  title={Optimal transport minimization: Crowd localization on density maps for semi-supervised counting},
  author={Lin, Wei and Chan, Antoni B},
  booktitle={2023 IEEE/CVF Conference on Computer Vision and Pattern Recognition (CVPR)},
  pages={21663--21673},
  year={2023},
  organization={IEEE}
}

@inproceedings{wan2021generalized,
  title={A generalized loss function for crowd counting and localization},
  author={Wan, Jia and Liu, Ziquan and Chan, Antoni B},
  booktitle={2021 IEEE/CVF conference on computer vision and pattern recognition (CVPR)},
  pages={1974--1983},
  year={2021},
  organization={IEEE}
}

@misc{qwen38,
    title  = {{Qwen3.8-Max}: A New Bar for Coding and Cowork},
    author = {{Qwen Team}},
    year   = {2026},
    month  = {August},
    url    = {https://qwen.ai/blog?id=qwen3.8}
}

@article{d2026recount,
  title={RECOUNT: Reference-guided Counting with Synthetic Visual Exemplars},
  author={D'Alessandro, Adriano and Mahdavi-Amiri, Ali and Hamarneh, Ghassan},
  journal={arXiv preprint arXiv:2608.20621},
  year={2026}
}

@inproceedings{
podell2024sdxl,
title={{SDXL}: Improving Latent Diffusion Models for High-Resolution Image Synthesis},
author={Dustin Podell and Zion English and Kyle Lacey and Andreas Blattmann and Tim Dockhorn and Jonas M{\"u}ller and Joe Penna and Robin Rombach},
booktitle={The Twelfth International Conference on Learning Representations},
year={2024}
}

@inproceedings{
liu2023visual,
title={Visual Instruction Tuning},
author={Haotian Liu and Chunyuan Li and Qingyang Wu and Yong Jae Lee},
booktitle={Thirty-seventh Conference on Neural Information Processing Systems},
year={2023},
}

@article{yang2025qwen3,
  title={Qwen3 technical report},
  author={Yang, An and Li, Anfeng and Yang, Baosong and Zhang, Beichen and Hui, Binyuan and Zheng, Bo and Yu, Bowen and Gao, Chang and Huang, Chengen and Lv, Chenxu and others},
  journal={arXiv preprint arXiv:2505.09388},
  year={2025}
}

@inproceedings{wang2024vision,
  title={Vision transformer off-the-shelf: A surprising baseline for few-shot class-agnostic counting},
  author={Wang, Zhicheng and Xiao, Liwen and Cao, Zhiguo and Lu, Hao},
  booktitle={Proceedings of the AAAI Conference on Artificial Intelligence},
  year={2024}
}

@article{wang2025internvl3,
  title={Internvl3. 5: Advancing open-source multimodal models in versatility, reasoning, and efficiency},
  author={Wang, Weiyun and Gao, Zhangwei and Gu, Lixin and Pu, Hengjun and Cui, Long and Wei, Xingguang and Liu, Zhaoyang and Jing, Linglin and Ye, Shenglong and Shao, Jie and others},
  journal={arXiv preprint arXiv:2508.18265},
  year={2025}
}

@inproceedings{
hu2022lora,
title={Lo{RA}: Low-Rank Adaptation of Large Language Models},
author={Edward J Hu and yelong shen and Phillip Wallis and Zeyuan Allen-Zhu and Yuanzhi Li and Shean Wang and Lu Wang and Weizhu Chen},
booktitle={International Conference on Learning Representations},
year={2022},
url={https://openreview.net/forum?id=nZeVKeeFYf9}
}
}

\clearpage
\appendix

\section{What Layout Components Matter Most?}
Table~\ref{tab:layout_ablation} explores the contribution of each layout component. The class name is always present, so every row is a request the counter can act on, and what varies is how much the layout says beyond the name. On FSC-147 the whole table spans $0.84$ MAE, while on LOOKALIKES it spans $4.03$ and the region alone is worth $2.85$. This is the expected result, as FSC-147 annotates only one category per image, so is minimally impacted by swapping out layout components. Whereas, LOOKALIKES puts multiple visually similar categories in an image, and so negative rows become load bearing, worth only $0.3$ on FSC-147 and between $1.16$ and $2.30$ on LOOKALIKES. Instance rows are the exception and run the other way, costing $1.06$ on LOOKALIKES when added to a declared region, which suggests that pointing at examples of a category is a weaker instruction than bounding where it is and naming what it is not.

\begin{table}[b]
\centering\small
\setlength{\tabcolsep}{4pt}
\caption{\textbf{What the layout contributes.} Layouts are composed
from a combination of region rows, instance rows and negative rows. We explore the contribution of each row type. \emph{Full} is a region covering the whole image, \emph{declared} is the region the upsampler names, \emph{none} omits the region row. \checkmark = the presence of either the instance row or negative row. Single forward pass, no tiling, so the layout is isolated from the inference-time blend.}
\label{tab:layout_ablation}
\begin{tabular}{@{}lcc cc cc@{}}
\toprule
& & & \multicolumn{2}{c}{FSC-147} & \multicolumn{2}{c}{LOOKALIKES} \\
\cmidrule(lr){4-5}\cmidrule(lr){6-7}
Region & Inst. & Neg. & MAE & RMSE & MAE & RMSE \\
\midrule
full     &            &            & 10.24 & 93.00 & 10.37 & 22.18 \\
full     &            & \checkmark & 10.15 & 94.15 &  8.22 & 21.40 \\
full     & \checkmark &            &  9.94 & 94.03 & 10.40 & 23.30 \\
full     & \checkmark & \checkmark &  9.92 & 93.97 &  8.10 & 19.88 \\
\midrule
declared &            &            & 10.68 & 93.39 &  7.53 & 19.56 \\
declared &            & \checkmark & 10.38 & 94.40 & \textbf{6.37} & 18.72 \\
declared & \checkmark &            &  9.96 & 93.79 &  8.58 & 19.90 \\
declared & \checkmark & \checkmark &  9.92 & 93.35 &  6.85 & \textbf{17.39} \\
\midrule
none     & \checkmark &            &  9.91 & 93.81 &  9.08 & 20.70 \\
none     & \checkmark & \checkmark & \textbf{9.83} & \textbf{93.12} &  7.30 & 18.31 \\
\bottomrule
\end{tabular}
\end{table}

\section{Swapping the Upsampler}
Table~\ref{tab:upsampler_substitution} replaces that upsampler at inference with two others and changes nothing else. Both are read without retraining. The deployed Qwen3.8 upsampler~\cite{qwen38} remains the
best of the three, with InternVL3.5~\cite{wang2025internvl3} at $10.01$ and the FLUX.3 upsampler~
\cite{flux-3-image-klein-2026} at $10.52$. However, the Qwen3.8 upsampler also has the most parameters. 

The gap between them is not primarily related to layout quality. An upsampler supplies two things, a layout and a count, and the count enters the prediction twice, once as an estimate in the median and once as the signal that decides when to tile. Holding the count fixed and varying only the layout, the FLUX.3 upsampler matches the deployed one, which says the counter reads a layout it was not trained against.

\begin{table}[t]
\centering\small
\caption{\textbf{Substituting the prompt upsampler.} One counter, trained once, provided layouts from three upsamplers under the deployed recipe. Each upsampler supplies both its layout and its own count. Where an upsampler returns no layout the counter falls back to the full frame and the class name, which is available without any upsampler; where it returns no count, the untiled prediction takes its place in the median. \emph{Fallback} reports how often the first case applies.}
\label{tab:upsampler_substitution}
\begin{tabular}{ll ccc}
\toprule
Upsampler & Params & MAE$\downarrow$ & RMSE$\downarrow$ & Fallback \\
\midrule 
Qwen3.8 & 27B      & 7.06 & 19.38 & 0 \\
InternVL3.5 & 8B   & 10.01 & 32.36 & 2 \\
FLUX.3 upsampler & 9B & 10.52 & 95.10 & 69 \\
\bottomrule
\end{tabular}
\end{table}

\section{Related Work}
\label{sec:related}

\paragraph{Zero-shot counting.} Zero-shot counting was formalised by~\citet{Xu_2023_CVPR}, which names the target category with a text prompt. Closest to us are the methods that seek exemplar precision while staying text-guided. ZSC~\citep{Xu_2023_CVPR}, CountSE~\citep{liu2025countse} and CountGD++~\citep{Amini_Naieni_2026_CVPR} mine pseudo-exemplars from the query image under text guidance. We read this as a special case of a larger question. Selecting which features are the target presumes an answer to what is in the scene, where it is and what else resembles it, and a model that answers that question supplies the pseudo-exemplars as a by-product. We therefore keep text as the channel and widen what it carries, from a class name to a layout that states the region, the examples, the distractors and what counts as one object.

\paragraph{Counting with synthetic data.} Early counters used 3D renderings such as GCC~\citep{wang2019learning}, while recent work uses diffusion models. FiGO~\citep{dalessandro2025justsaytheword} generates data to adapt a segmentation model to new categories at test time, AFreeCA~\citep{d2024afreeca} extracts a coarse quantity signal from generated images to supervise counting without labels, and CountGD++~\citep{Amini_Naieni_2026_CVPR} generates a canonical object to serve as a pseudo-exemplar. These generate an exemplar or a category. We generate a counterfactual: one scene rendered twice with
the objects exchanged for a sibling category, composited so that a single
image admits several valid density maps. The purpose differs accordingly.
Theirs is to describe the target better, ours is to make a layout impossible
to ignore, since counting data that pairs each image with one valid answer
teaches a counter to disregard the prompt.

\section{Implementation Details}
\label{sec:supp_impl}

\subsection{Counter}

The counter is FLUX.3 Image Klein~\cite{flux-3-image-klein-2026}. The backbone is frozen and adapted with a low-rank adapter~\cite{hu2022lora} of rank $128$ and $\alpha=128$, zero dropout, Gaussian initialisation, applied to the attention projections, the MLP input and output projections, the text embedder, the image and image-condition embedders, and the output projection. There are $263$ adapter sites and $326.7$M trainable parameters.

Training runs for $6{,}000$ optimiser steps at an effective batch size of $2$, at a resolution of $768\times576$. The learning rate is $2\times10^{-4}$ under a three-phase linear schedule with a
$200$-step warmup from $0.01\times$ the base rate, then linear decay to $0.1\times$ the base rate at the final step. Gradients are clipped to a global norm of $1.0$.

\subsection{Target Encoding}

Point annotations are rendered into a density map with a Gaussian kernel whose width is adaptive to local spacing, using $k=2$ nearest neighbours, $\beta=0.30$ and an upper bound $\sigma_{\text{hi}}=32$ pixels. The rendered map integrates to the object count exactly.

The map is encoded for the latent space with a logarithmic transform at peak $0.2$ and $\varepsilon=10^{-6}$, mapped to $[-1,1]$, and the count is recovered by inverting that transform before summing.

\subsection{Losses}

Three terms, all at weight $1.0$ in the deployed run.

\paragraph{Latent regression} between the predicted density latent and the ground-truth map passed through the frozen encoder.

\paragraph{Generalized Loss} is computed in pixel space on the decoded prediction against the point set, at stride $8$, with blur $8.0$ and reach $48.0$ pixels, $\tau=0.1$ for the two marginal-penalty terms and the gating, and normalised by count.

\paragraph{Negative mass} scores predicted mass falling within $12$ pixels of a negative point, at stride $4$, capped at $2{,}048$ negative points per image. It is a \emph{fraction} of predicted mass, so it redistributes rather than competing for the total, and it applies only to records carrying a second annotated category.

\subsection{Training Data}

Six streams are sampled jointly: FSC-147 train, a dense-reweighted FSC-147
stream, real seam composites, synthetic composites, sibling composites, and
attribute composites. A stream defined but not listed in the training
mapping contributes nothing, so the mapping is asserted at startup.

\paragraph{Composite preparation.} A scene is generated for one category
with a text-to-image model using category names from a prior work~\cite{d2026recount}, and an image-to-image model replaces the objects and only the objects with those of a sibling category, and the two renderings are composited along a straight cut. Each record stores the cut normal, offset, margin and whether the cut crosses the object field. Both halves are pseudo-labelled independently by an off-the-shelf counting model, which is reliable because each half contains a single category by construction. 

\subsection{Prompt Construction}

Training layouts are generated per record with the following randomization,
at the values the deployed run set:

\begin{itemize}
  \item a real upsampler caption replaces the generated layout with probability $0.35$;
  \item a declared negative is dropped with probability $0.21$;
  \item the target region is inflated with probability $0.30$;
  \item rows are shuffled on every sample, and negatives are enumerated in prose on every sample.
\end{itemize}

\subsection{Inference}

The image is encoded once and the density latent obtained in a single forward pass at a fixed timestep of $0.499$. There is no sampler, schedule or seed, so a prediction is deterministic. The latent is decoded and the count read as the sum of the map.

Tiling uses a $3\times3$ grid with $0.3$ overlap, windows taken on the declared region with box-aspect geometry, skipping a tile whose predicted occupancy falls below $0.05$ or whose mass falls below $0.01$, and passing non-empty tiles through a generative upscaler. Exemplar and exclusion rows are carried into each tile prompt.

The upsampler's own count decides when to tile. We take the median of the untiled count, the tiled count and the upsampler count; above an upsampler count of $400$ we move to $5\times5$, and above $1{,}000$ to a raw $9\times9$. When an upsampler returns no count, the untiled prediction is
substituted for it, which reduces the median to the untiled estimate.

\end{document}